\documentclass{IOS-Book-Article}

\usepackage{mathptmx}

\usepackage[utf8]{inputenc} 
\RequirePackage{hyperref}
\RequirePackage{hypernat}

\usepackage{graphicx}
\usepackage{url}
\usepackage{amssymb}
\usepackage[dvipsnames]{xcolor}
\usepackage{enumerate}
\usepackage{algpseudocode, algorithm}

\usepackage{caption}
\usepackage{subcaption}
\usepackage{multirow}
\usepackage{amsmath}
\usepackage{adjustbox}
\usepackage{makecell}
\usepackage{xcolor}

\def\hb{\hbox to 10.7 cm{}}

\begin{document}

\pagestyle{headings}
\def\thepage{}

\begin{frontmatter} 

\title{Neuro-symbolic Architectures for Context Understanding}

\markboth{}{January 2020\hb}
\footnotetext{All authors contributed equally to the chapter.}
\author[A]{\fnms{Alessandro} \snm{Oltramari}}
\author[A,B]{\fnms{Jonathan} \snm{Francis}}
\author[A]{\fnms{Cory} \snm{Henson}}
\author[B]{\fnms{Kaixin} \snm{Ma}}\footnote{Work done during an internship at Bosch Research \& Technology Center Pittsburgh.}
\author[C]{\fnms{Ruwan} \snm{Wickramarachchi}}\footnotemark[1]

\address[A]{Intelligent IoT, Bosch Research and Technology Center (Pittsburgh, PA, USA)}
\address[B]{Language Technologies Institute, School of Computer Science, Carnegie Mellon University (Pittsburgh, PA, USA)}
\address[C]{Artificial Intelligence Institute, University of South Carolina (Columbia, SC, USA)}

\begin{abstract}
Computational context understanding refers to an agent's ability to fuse disparate sources of information for decision-making and is, therefore, generally regarded as a prerequisite for sophisticated machine reasoning capabilities, such as in artificial intelligence (AI). \textit{Data-driven} and \textit{knowledge-driven} methods are two classical techniques in the pursuit of such machine sense-making capability. However, while data-driven methods seek to model the statistical regularities of events by making observations in the real-world, they remain difficult to interpret and they lack mechanisms for naturally incorporating external knowledge. Conversely, knowledge-driven methods, combine structured knowledge bases, perform symbolic reasoning based on axiomatic principles, and are more interpretable in their inferential processing; however, they often lack the ability to estimate the statistical salience of an inference. To combat these issues, we propose the use of \textit{hybrid} AI methodology as a general framework for combining the strengths of both approaches. Specifically, we inherit the concept of \textit{neuro-symbolism} as a way of using knowledge-bases to guide the learning progress of deep neural networks. We further ground our discussion in two applications of neuro-symbolism and, in both cases, show that our systems maintain interpretability while achieving comparable performance, relative to the state-of-the-art.
\end{abstract}

\end{frontmatter}
\markboth{January 2020\hb}{January 2020\hb}

\section{Explainability through Context Understanding}

Context understanding is a natural property of human cognition, that supports our decision-making capabilities in complex sensory environments. Humans are capable of fusing information from a variety of modalities|e.g., auditory, visual|in order to perform different tasks, ranging from the operation of a motor vehicle to the generation of logical inferences based on commonsense. Allen Newell and Herbert Simon described this \textit{sense-making capability} in their theory of cognition \cite{newell1994unified, newell1972human}: through sensory stimuli, humans accumulate experiences, generalize, and reason over them, storing the resulting knowledge in memory; the dynamic combination of live experience and distilled knowledge during task-execution, enables humans to make time-effective decisions and evaluate how good or bad a decision was by factoring in external feedback. \\
Endowing machines with this sense-making capability has been one of the long-standing goals of Artificial Intelligence (AI) practice and research, both in industry and academia. \textit{Data-driven} and \textit{knowledge-driven} methods are two classical techniques in the pursuit of such machine sense-making capability. Sense-making is not only a key for improving machine autonomy, but is a precondition for enabling seamless interaction with humans. Humans communicate effectively with each other, thanks to their shared mental models of the physical world and social context~\cite{johnson1995mental}. These models foster reciprocal trust by making contextual knowledge transparent; they are also crucial for explaining how decision-making unfolds. In a similar fashion, we can assert that `explainable AI' is a byproduct or an {\it affordance} of computational context understanding and is predicated on the extent to which humans can introspect the decision processes that enable machine sense-making \cite{lawless2019artificial}.

\section{Context Understanding through Neuro-symbolism} 

From the definitions of `explainable AI' and `context understanding,' in the previous section, we can derive the following corollary: \begin{quote}
    {\it The explainability of AI algorithms is related to how context is processed, computationally, based on the machine's perceptual capabilities and on the external knowledge resources that are available.}
\end{quote} 
Along this direction, the remainder of this chapter explores two concrete scenarios of context understanding, realized by \textit{neuro-symbolic architectures}|i.e., hybrid AI frameworks that instruct machine perception (based on deep neural networks) with knowledge graphs\footnote{Although inspired by the human capability of fusing perception and knowledge, the neuro-symbolic architectures we illustrate in this chapter do not commit on replicating the mechanisms of human context understanding|this being the scientific tenet of cognitive architecture research (see, e.g., \cite{laird2017standard, anderson2009can}).}. These examples were chosen to illustrate the general applicability of neuro-symbolism and its relevance to contemporary research problems.\\
Specifically, section \ref{application-1} considers context understanding for autonomous vehicles: we describe how a knowledge graph can be built from a dataset of urban driving situations and how this knowledge graph can be translated into a continuous vector-space representation. This embedding space can be used to estimate the semantic similarity of visual scenes by using neural networks as powerful, non-linear function approximators. Here, models may be trained to make danger assessments of the visual scene and, if necessary, transfer control to the human in complex scenarios. The ability to make this assessment is an important capability for autonomous vehicles, when we consider the negative ramifications for a machine to remain invariant to changing weather conditions, anomalous behavior of dynamic obstacles on the road (e.g., other vehicles, pedestrians), varied lighting conditions, and other challenging circumstances. We suggest neuro-symbolic fusion as one solution and, indeed, our results show that our embedding space preserves the semantic properties of the conceptual elements that make up visual scenes. 

%
In section \ref{application-2}, we describe context understanding for language tasks.
Here, models are supplied with three separate modalities: external commonsense knowledge, unstructured textual context, and a series of answer candidates. In this task, models are tested on their ability to fuse together these disparate sources of information for making the appropriate logical inferences. We designed methods to extract adequate semantic structures (i.e., triples) from two comprehensive commonsense knowledge graphs, \texttt{ConceptNet} \cite{liu2004conceptnet} and \texttt{Atomic} \cite{sap2019atomic}, and to inject this external context into language models. In general, open-domain linguistic context is useful for different tasks in Natural Language Processing (NLP), including: information-extraction, text-classification, extractive and abstractive summarization, and question-answering (QA). For ease of quantitative evaluation, we consider a QA task in section \ref{application-2}. In particular, the task is to select the correct answer from a pool of candidates, given a question that specifically requires commonsense to resolve. For example, the question, \textit{If electrical equipment won't power on, what connection should be checked?} is associated with `company', `airport', `telephone network', `wires', and `freeway'(where `wires' is the correct answer choice). We demonstrate that our proposed hybrid architecture out-performs the state-of-the-art neural approaches that do not utilize structured commonsense knowledge bases. 
Furthermore, we discuss how our approach maintains explainability in the model's decision-making process: the model has the joint task of learning an attention distribution over the commonsense knowledge context which, in turn, depends on the knowledge triples that were conceptually most salient for selecting the correct answer candidate, downstream. 
Fundamentally, the goal of this project is to make human interaction with chatbots and personal assistants more robust. For this to happen, it is crucial to equip intelligent agents with a shared understanding of general contexts, i.e., commonsense. Conventionally, machine commonsense had been computationally articulated using symbolic languages|\texttt{Cyc} being one of the most prominent outcomes of this approach \cite{matuszek2006introduction}. However, symbolic commonsense representations are neither scalable nor comprehensive, as they depend heavily on the knowledge engineering experts that encode them. In this regard, the advent of deep learning and, in particular, the possibility of fusing symbolic knowledge into sub-symbolic (neural) layers, has recently led to a revival of this AI research topic.

\section{Applications of Neuro-symbolism}
\label{applications}


\subsection{Application I: Learning a Knowledge Graph Embedding Space for Context Understanding in Automotive Driving Scenes}
\label{application-1}

\subsubsection{Introduction}

Recently, there has been a significant increase in the investment for autonomous driving (AD) research and development, with the goal of achieving full autonomy in the next few years. Realizing this vision requires robust ML/AI algorithms that are trained on massive amounts of data. Thousands of cars, equipped with various types of sensors (e.g., LIDAR, RGB, RADAR), are now deployed around the world to collect this heterogeneous data from real-world driving scenes. The primary objective for AD is to use these data to optimize the vehicle's \textit{perception pipeline} on such tasks as: 3D object detection, obstacle tracking, object trajectory forecasting, and learning an ideal driving policy. Fundamental to all of these tasks will be the vehicle's context understanding capability, which requires knowledge of the time, location, detected objects, participating events, weather, and various other aspects of a driving scene. Even though state-of-the-art AI technologies are used for this purpose, their current effectiveness and scalability are insufficient to achieve full autonomy. Humans naturally exhibit context understanding behind the wheel, where the decisions we make are the result of a continuous evaluation of perceptual cues combined with background knowledge. For instance, human drivers generally know which area of a neighborhood might have icy road conditions on a frigid winter day, where flooding is more frequent after a heavy rainfall, which streets are more likely to have kids playing after school, and which intersections have poor lighting. Currently, this type of common knowledge is not being used to assist self-driving cars and, due to the sample-inefficiency of current ML/AI algorithms, vehicle models cannot effectively learn these phenomena through statistical observation alone. 

On March 18, 2018, Elaine Herzberg’s death was reported as the first fatality incurred from a collision with an autonomous vehicle\footnote{https://www.nytimes.com/2018/03/19/technology/uber-driverless-fatality.html}. An investigation into the collision, conducted by The National Transportation Safety Board (NTSB), remarks on the shortcomings of current AD and context understanding technologies. Specifically, NTSB found that the autonomous vehicle incorrectly classified Herzberg as an unknown object, a vehicle, and then a bicycle within the complex scene as she walked across the road. Further investigation revealed that the system design did not include consideration for pedestrians walking outside of a crosswalk, or jaywalking \cite{Inadequ56:online}. Simply put, the current AD technology lacks fundamental understanding of the characteristics of objects and events within common scenes; this suggests that more research is required in order to achieve the vision of autonomous driving. 

Knowledge Graphs (KGs) have been successfully used to manage heterogeneous data within various domains. They are able to integrate and structure data and metadata from multiple modalities into a unified semantic representation, encoded as a graph. More recently, KGs are being translated into latent vector space representations, known as Knowledge Graph Embeddings (KGEs), that have been shown to improve the performance of machine learning models when applied to certain downstream tasks, such as classification \cite{chen2017multilingual,wang2019multi}. Given a KG as a set of triples, KGE algorithms learn to create a latent representation of the KG entities and relations as continuous KGE vectors. This encoding allows KGEs to be easily manipulated and integrated with machine learning algorithms. Motivated by the shortcomings of current context understanding technologies, along with the promising outcomes of KGEs, our research focuses on the generation and evaluation of KGEs on AD data. Before directly applying KGEs on critical AD applications, however, we evaluate the intrinsic quality of KGEs across multiple metrics and KGE algorithms \cite{wickramarachchi2020make}. Additionally, we present an early investigation of using KGEs for a selected use-case from the AD domain. 

\subsubsection{Scene Knowledge Graphs}

\textbf{\\Dataset.} To promote and enable further research on autonomous driving, several benchmark datasets have been made publicly available by companies in this domain \cite{laflamme2019driving}. \texttt{NuScenes} is a benchmark dataset of multimodal vehicular data, recently released by Aptiv \cite{nuscenes2019} and used for our experiments. \texttt{NuScenes} consists of a collection of 20-second driving scenes, with $\sim$40 sub-scenes sampled per driving scene (i.e., one every 0.5 seconds). In total, \texttt{NuScenes} includes 850 driving scenes and 34,149 sub-scenes. Each sub-scene is annotated with detected objects and events, each defined within a taxonomy of 23 object/event categories.

\begin{figure}[!h]
    \centering
    \includegraphics[scale=0.75]{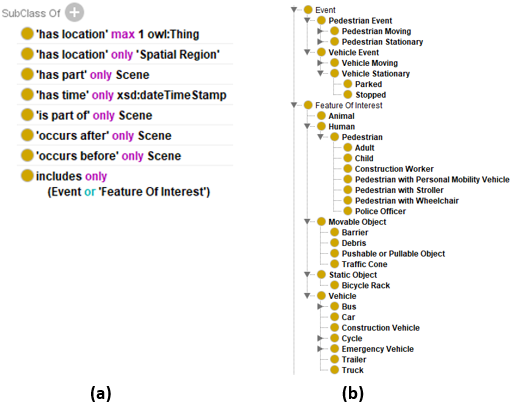}
    \caption{Scene Ontology: (a) formal definition of a \textit{Scene}, and (b) a subset of \textit{Features-of-Interests} and \textit{events} defined within a taxonomy.}
    \label{fig:scene-ontology}
\end{figure}

\textbf{\\Scene Ontology.} In autonomous driving, a scene is defined as an observable volume of time and space \cite{henson2019iswc}. On the road, a vehicle may encounter many different situations|such as merging onto a divided highway, stopping at a traffic light, and overtaking another vehicle|all of which are considered as common driving scenes. A scene encapsulates all relevant information about a particular situation, including data from vehicular sensors, objects, events, time and location. A scene can also be divided into a sequence of sub-scenes. As an example, a 20-second drive consisting primarily of the vehicle merging into a highway could be considered as a scene. In addition, all the different situations the vehicle encounters within these 20 seconds can also be represented as (sub-)scenes. In this case, a scene may be associated with a time interval and spatial region while a sub-scene may be associated with a specific timestamp and a set of spatial coordinates. This semantic representation of a scene is formally defined in the \texttt{Scene Ontology} (see figure \ref{fig:scene-ontology}(a), depicted in Protege\footnote{https://protege.stanford.edu/}). To enable the generation of a KG from the data within \texttt{NuScenes}, the \texttt{Scene Ontology} is extended to include all the concepts (i.e., objects and event categories) found in the \texttt{NuScenes} dataset.

\textbf{\\Generating Knowledge Graphs.} The Scene Ontology identifies \textit{events} and \textit{features-of-interests} (FoIs) as top-level concepts. An \textit{event} or a \textit{FoI} may be associated with a \textit{Scene} via the \textit{includes} relation. \textit{FoIs} are associated with \textit{events} through the \textit{isParticipantOf} relation. Figure \ref{fig:scene-ontology}(b) shows a subset of the \textit{FoIs} and \textit{events} defined by the \texttt{Scene Ontology}. In generating the scenes' KG, each scene and sub-scene found in \texttt{NuScenes} is annotated using the \texttt{Scene Ontology}. Table \ref{tab:kg-stats} shows some basic statistics of the generated KG.  

\begin{table}[htbp]
\centering
\begin{tabular}{l|l}
\Xhline{2\arrayrulewidth}
\# of triples & 5.95M \\
\# of entities & 2.11M \\
\# of relations & 11 \\ \Xhline{2\arrayrulewidth}

\end{tabular} 
\caption{Statistics of the scene KG generated from the \texttt{NuScenes} dataset}
\label{tab:kg-stats}
\end{table}

\subsubsection{Knowledge Graph Embeddings}
\textbf{\\KGE Algorithms.} KGE algorithms enable the ability to easily feed knowledge into ML algorithms and improve the performance of learning tasks, by translating the knowledge contained in knowledge graphs into latent vector space representation of KGEs \cite{myklebust2019knowledge}. To select candidate KGE algorithms for our evaluation, we referred to the classification of KGE algorithms provided by Wang et al. \cite{wang2017knowledge}. In this work, KGE algorithms are classified into two primary categories: (1) Transitional distance-based algorithms and (2) Semantic matching-based models. Transitional distance-based algorithms define the scoring function of the model as a distance-based measure, while semantic matching-based algorithms define it as a similarity measure. Here, entity and relation vectors interact via addition and subtraction in the case of Transitional distance-based models; in semantic matching-based models, the interaction between entity and relation vectors is captured by multiplicative score functions \cite{sharma2018towards}.

Initially, for our study we had selected one algorithm from each class: TransE \cite{bordes2013translating} to represent the transitional distance-based algorithms and RESCAL \cite{nickel2011three} to represent the semantic matching-based algorithms. However, after experimentation, RESCAL did not scale well for handling large KGs in our experiments. Therefore, we also included HolE \cite{nickel2016holographic}|an efficient successor of RESCAL|in the evaluation. A brief summary of each algorithm is provided for each model, below: \\

\noindent
\textit{TransE:} the TransE model is often considered to be the most-representative of the class of transitional distance-based algorithms \cite{wang2017knowledge}. Given a triple \textit{(h, r, t)} from the KG, TransE encodes \textit{h}, \textit{r} and \textit{t} as vectors, with \textit{r} represented as a transition vector from \textit{h} to \textit{t}: \textit {$\mathbf{h} + \mathbf{r} \approx \mathbf{t}$}. Since both entities and relations are represented as vectors, TransE is one of the most efficient KGE algorithms, with $\mathcal{O}(n d + m d)$ space complexity and $\mathcal{O}(n_t d)$ time complexity ($n_t$ is the number of training triples).\\

\noindent
\textit{RESCAL:} RESCAL is capable of generating an expressive knowledge graph embedding space, due to its ability to capture complex patterns over multiple hops in the KG. RESCAL encodes relations as matrices and captures the interaction between entities and relations using a bi-linear scoring function. Though the use of a matrix to encode each relation yields improved expressivity, it also limits RESCAL’s ability to scale with large KGs. It has $\mathcal{O}(n d + m d^2)$ space complexity and $\mathcal{O}(n_t d^2)$ time complexity.\\

\noindent
\textit{HolE:} HoLE is a more efficient successor of RESCAL, addressing its space and time complexity issues, by encoding relations as vectors without sacrificing the expressivity of the model. By using circular correlation operation \cite{nickel2016holographic}, it captures the pairwise interaction of entities as composable vectors. This optimization yields $\mathcal{O} (n d + m d)$ space complexity and $\mathcal{O}(n_t d \log d)$ time complexity.

\begin{figure}[h]
    \centering
    \includegraphics[scale=0.33]{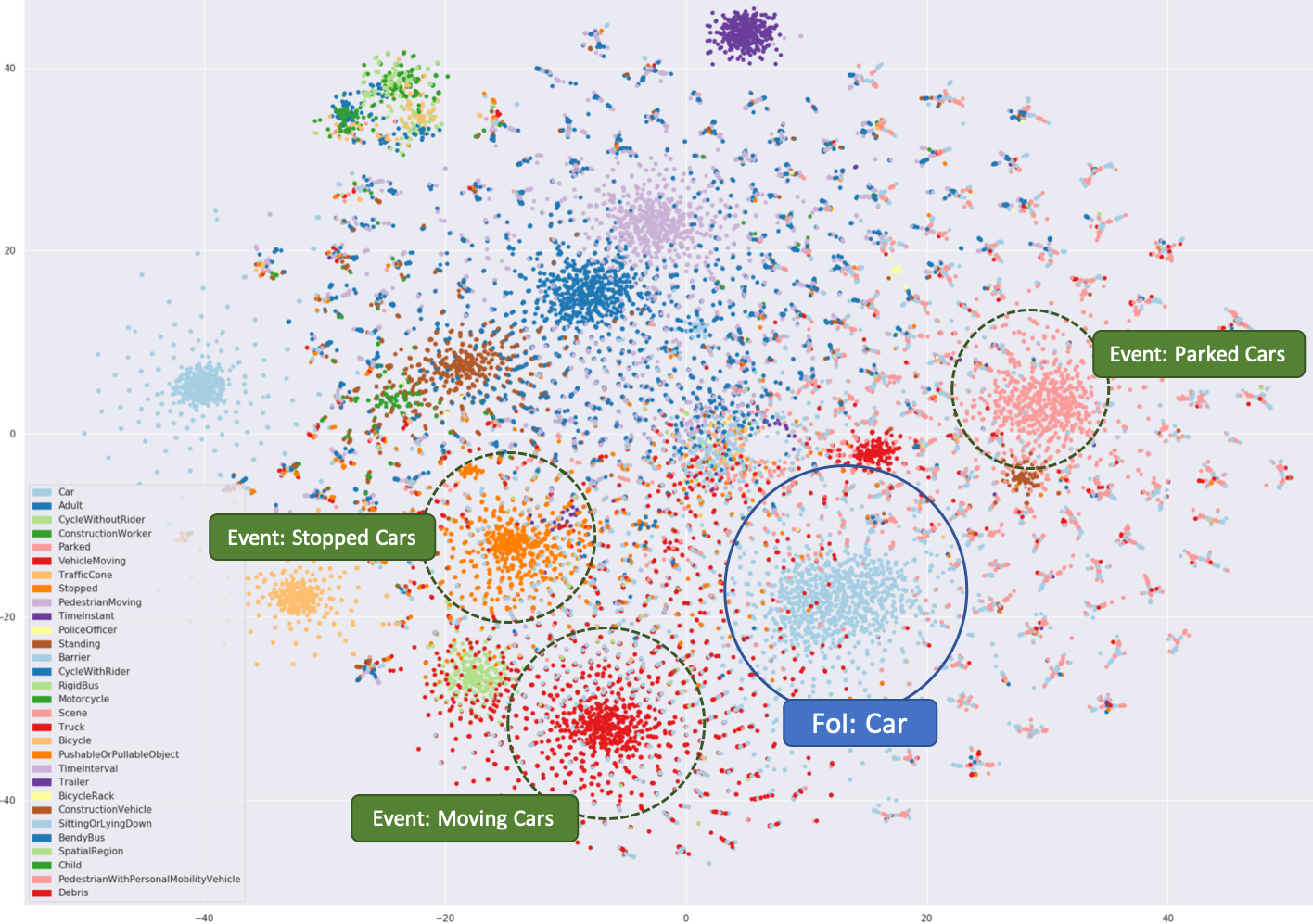}
    \caption{2D visualizations of KGEs of \texttt{NuScenes} instances generated from the TransE algorithm.}
    \label{fig:fois-events}
\end{figure}

\textbf {\\Visualizing KGEs.}
In order to visualize the generated KGE, a ``mini" KG from the \texttt{NuScenes-mini} dataset was created. Specifically, 10 scenes were selected (along with their sub-scenes) to generate the KG, and the TransE algorithm was used to learn the embeddings. When training the KGEs, we chose the dimension of the vectors to be 100. To visualize the embeddings in 2-dimensional (2D) space, the dimensions are reduced using the t-Distributed Stochastic Neighbor Embedding (t-SNE) \cite{maaten2008visualizing} projection. Figure \ref{fig:fois-events}(a) shows the resulting embeddings of the \texttt{NuScenes} dataset. To denote interesting patterns that manifest in the embeddings, instances of \textit{Car} (a \textit{FoI}) and the \textit{events} in which they participate are highlighted. In this image, events such as \textit{parked car}, \textit{moving car}, and \textit{stopped car} are clustered around entities of type \textit{Car}. This shows that the \textit{isParticipantOf} relations defined in the KG are maintained within the KG embeddings.

\vspace{-2pt}
\subsubsection{Intrinsic Evaluation}
Here, we deviate slightly from the prior work in evaluating KGE algorithms, which evaluate KGEs based downstream task performance. Instead, we focus on an evaluation that uses only metrics that quantify the \textit{intrinsic} quality of KGEs \cite{alashargi2019metrics}: categorization measure, coherence measure, and semantic transition distance. Categorization measures how well instances of the same type cluster together. To quantify this quality, all vectors of the same type are averaged together and the cosine similarity is computed between the averaged vector and the typed class. The Coherence measure quantifies the proportion of neighboring entities that are of the same type; the evaluation framework proposes that, if a set of entities are typed by the class, those entities should form a cluster in the embedding space with the typed class as the centroid. Adapted from the word embedding literature, \textit{Semantic Transitional Distance} captures the relational semantics of the KGE: if a triple $(h,r,t)$ is correctly represented in the embedding space, the transition distance between the vectors representing $(\mathbf{h+r})$ should be close to $\mathbf{t}$. This is quantified by computing the cosine similarity between $(\mathbf{h+r})$ and $\mathbf{t}$. \\

\begin{figure}[h]
    \centering
    \includegraphics[scale=0.25]{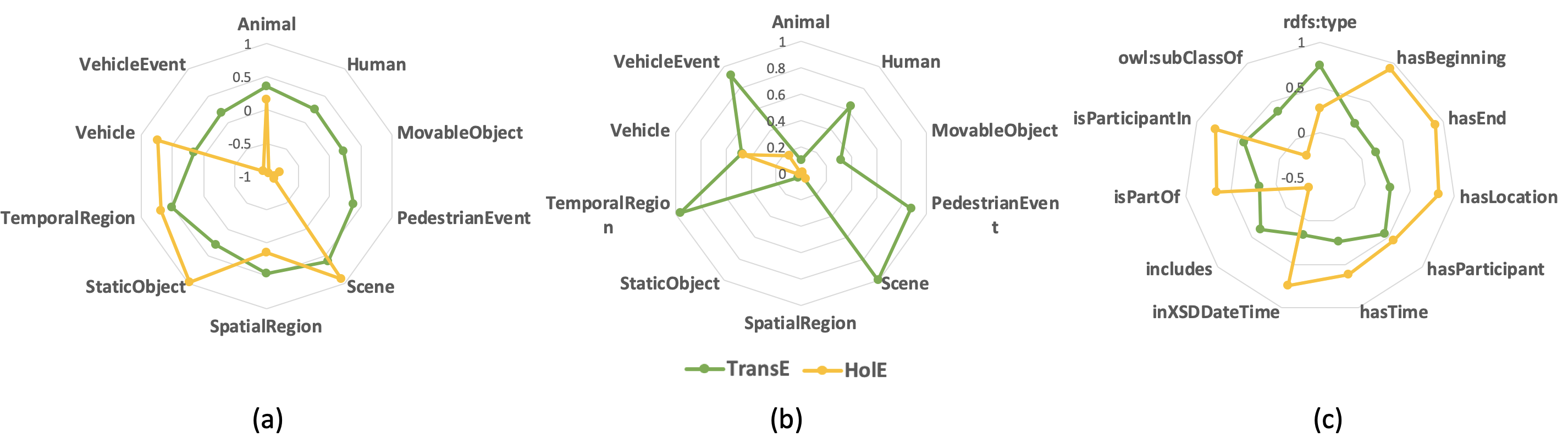}
    \caption{Evaluation results of the \texttt{NuScenes} dataset: (a) Categorization measure, (b) Coherence measure, and (c) Semantic transitional distance}
    \label{fig:nusc-base-eval}
\end{figure}

\noindent
\textbf{Results.} Evaluation results are reported with respect to each algorithm and metric. Figure \ref{fig:nusc-base-eval} shows the evaluation results of categorization measure, coherence measure, and semantic transitional distance|for each KGE algorithm. The \texttt{NuScenes} KG, generated from the \texttt{NuScenes-trainval} dataset, is large in terms of both the number of triples and number of entities (see Table \ref{tab:kg-stats}). Hence, RESCAL did not scale well to this dataset. For this reason, we only report the evaluation results for TransE and HolE. When considering the KGE algorithms, TransE's performance is consistently better across metrics, compared to HolE's performance. However, it is interesting to note that HolE significantly outperforms TransE for some classes/relations. When considering the evaluation metrics, it is evident that the categorization measure and semantic transitional distance are able to capture the quality of type semantics and relational semantics, respectively. The value of the coherence measure, however, is zero for HoLE in most cases and close to zero for TransE in some cases. In our experimental setting, the poor performance with respect to the coherence measure may suggest that it may not be a good metric for evaluating KGEs in the AD domain.

\subsubsection{A use-case from the AD domain}

\begin{figure}[!htbp]
    \centering
    \includegraphics[scale=0.4]{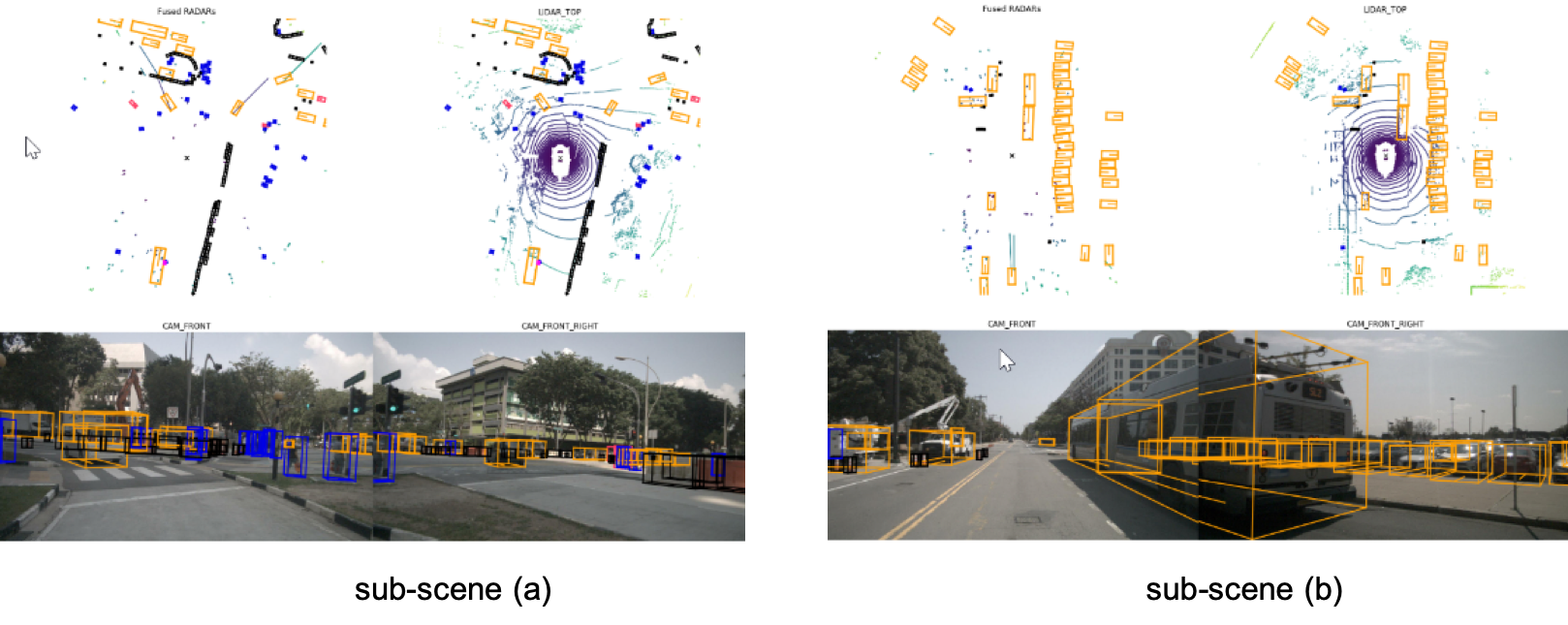}
    \caption{Results of scene similarity: Most similar sub-scenes computed using KGEs trained on \texttt{NuScenes} KG}
    \label{fig:scene-similarity}
\end{figure}

We report preliminary results from our investigation into using KGEs for a use-case in the AD domain. More specifically, we apply KGEs for computing scene similarity. In this case, the goal is to find (sub-)scenes that are characteristically similar, using the learned KGEs. Given a set of scene pairs, we choose the pair with the highest cosine similarity as the most similar. Figure \ref{fig:scene-similarity} shows an illustration of the two most similar sub-scenes, when the list of pairs include sub-scenes from different scenes. An interesting observation is that the black string of objects in sub-scene (a) are \textit{Barriers} (\textit{a Static Object}), and the orange string of objects in sub-scene (b) are \textit{Stopped Cars}. This example suggests that the KGE-based approach could identify sub-scenes that share similar characteristics even though the sub-scenes are visually dissimilar.

\subsubsection{Discussion}
We presented an investigation of using KGEs for AD context understanding, along with an evaluation of the intrinsic quality of KGEs. The evaluation suggests that KGEs are specifically able to capture the semantic properties of a scene knowledge graph (e.g., \textit{isParticipantOf} relation between objects and events). More generally, KGE algorithms are capable of translating semantic knowledge, such as type and relational semantics to KGEs. When considering the different KGE algorithms, we report that the transitional distance-based algorithm, TransE, shows consistent performance across multiple quantitative KGE-quality metrics. Our evaluation further suggests that some quality metrics currently in use, such as the coherence measure, may not be effective in measuring the quality of the type semantics from KGEs, in the AD domain. Finally, in applying the learned KGEs to a use-case of importance in the AD domain, we shed some light on the effectiveness of leveraging KGEs in capturing AD scene similarity. 



\subsection{Application II: Neural Question-Answering using Commonsense Knowledge Bases}
\label{application-2}
\subsubsection{Introduction}
Recently, many efforts have been made towards building challenging question-answering (QA) datasets that, by design, require models to synthesize external commonsense knowledge and leverage more sophisticated reasoning mechanisms  \cite{TACL1534,DBLP:journals/corr/abs-1810-12885,ostermann-etal-2018-semeval,zellers-etal-2018-swag,zellers-etal-2019-hellaswag}. Two directions of work that try to solve these tasks are: purely \textit{data-oriented} and purely \textit{knowledge-oriented} approaches. The data-oriented approaches generally propose to pre-train language models on large linguistic corpora, such that the model would implicitly acquire ``commonsense'' through its statistical observations. Indeed, large pre-trained language models have achieved promising performance on many commonsense reasoning benchmarks \cite{devlin-etal-2019-bert,radford2019language,yang2019xlnet,DBLP:journals/corr/abs-1907-11692}. The main downsides of this approach are that models are difficult to interpret and that they lack mechanisms for incorporating explicit commonsense knowledge. Conversely, purely knowledge-oriented approaches combine structured knowledge bases and perform symbolic reasoning, on the basis of axiomatic principles. Such models enjoy the property of interpretability, but often lack the ability to estimate the statistical salience of an inference, based on real-world observations. Hybrid models are those that attempt to fuse these two approaches, by extracting knowledge from structured knowledge bases and using the resulting information to guide the learning paradigm of statistical estimators, such as deep neural network models. 

Different ways of injecting knowledge into models have been introduced, such as attention-based gating mechanisms \cite{bauer-etal-2018-commonsense}, key-value memory mechanisms \cite{miller-etal-2016-key, mihaylov-frank-2018-knowledgeable}, extrinsic scoring functions \cite{DBLP:journals/corr/abs-1809-03568}, and graph convolution networks \cite{DBLP:journals/corr/KipfW16,Lin2019KagNetKG}. 
Our approach is to combine the powerful pre-trained language models with structured knowledge, and we extend previous approaches by taking a more fine-grained view of commonsense. The subtle differences across the various knowledge types have been discussed at length in AI by philosophers, computational linguists, and cognitive psychologists \cite{davis2014representations}. At the high level, we can identify \textit{declarative commonsense}, whose scope encompasses \text{factual knowledge}, e.g., `the sky is blue' and `Paris is in France'; \textit{taxonomic knowledge}, e.g., `football players are athletes' and `cats are mammals'; \textit{relational knowledge}, e.g., `the nose is part of the skull' and `handwriting requires a hand and a writing instrument'; \textit{procedural commonsense}, which includes prescriptive knowledge, e.g., `one needs an oven before baking cakes' and `the electricity should be off while the switch is being repaired' \cite{hobbs1987commonsense}; \textit{sentiment knowledge}, e.g., `rushing to the hospital makes people worried' and `being in vacation makes people relaxed'; and \textit{metaphorical knowledge} which includes idiomatic structures, e.g., `time flies' and `raining cats and dogs'. We believe that it is important to identify the most appropriate commonsense knowledge type required for specific tasks, in order to get better downstream performance. Once the knowledge type is identified, we can then select the appropriate knowledge base(s), the corresponding knowledge-extraction pipeline, and the suitable neural injection mechanisms.

In this work, we conduct a comparison study of different knowledge bases and knowledge-injection methods, on top of pre-trained neural language models; we evaluate model performance on a multiple-choice QA dataset, which explicitly requires commonsense reasoning. In particular, we used \texttt{ConceptNet} \cite{liu2004conceptnet} and the recently-introduced \texttt{ATOMIC} \cite{sap2019atomic} as our external knowledge resources, incorporating them in the neural computation pipeline using the \textit{Option Comparison Network} (OCN) model mechanism \cite{DBLP:journals/corr/abs-1903-03033}. We evaluate our models on the \texttt{CommonsenseQA} \cite{talmor-etal-2019-commonsenseqa} dataset; an example question from the \texttt{CommonsenseQA} task is shown in Table \ref{csqa-example}. Our experimental results and analysis suggest that attention-based injection is preferable for knowledge-injection and that the degree of domain overlap, between knowledge-base and dataset, is vital to model success.\footnote{From a terminological standpoint, `domain overlap' here should be interpreted as the overlap, between question types in the targeted datasets and types of commonsense represented in the knowledge bases under consideration.} 

\begin{table}[h]
\footnotesize
\begin{center}
\begin{tabular}{|l|}
\hline \bf Question: \\ 
A revolving door is convenient for two direction travel, but it also serves as a security measure at a what? \\
\textbf{Answer choices:}\\ 
A. Bank\textbf{*}; B. Library; C. Department Store; D. Mall; E. New York; \\
\hline
\end{tabular}
\end{center}
\caption{An example from the \texttt{CommonsenseQA} dataset; the asterisk (\textbf{*}) denotes the correct answer.}
\label{csqa-example}
\end{table}

\begin{figure*}
    \centering
    \includegraphics[scale=0.2]{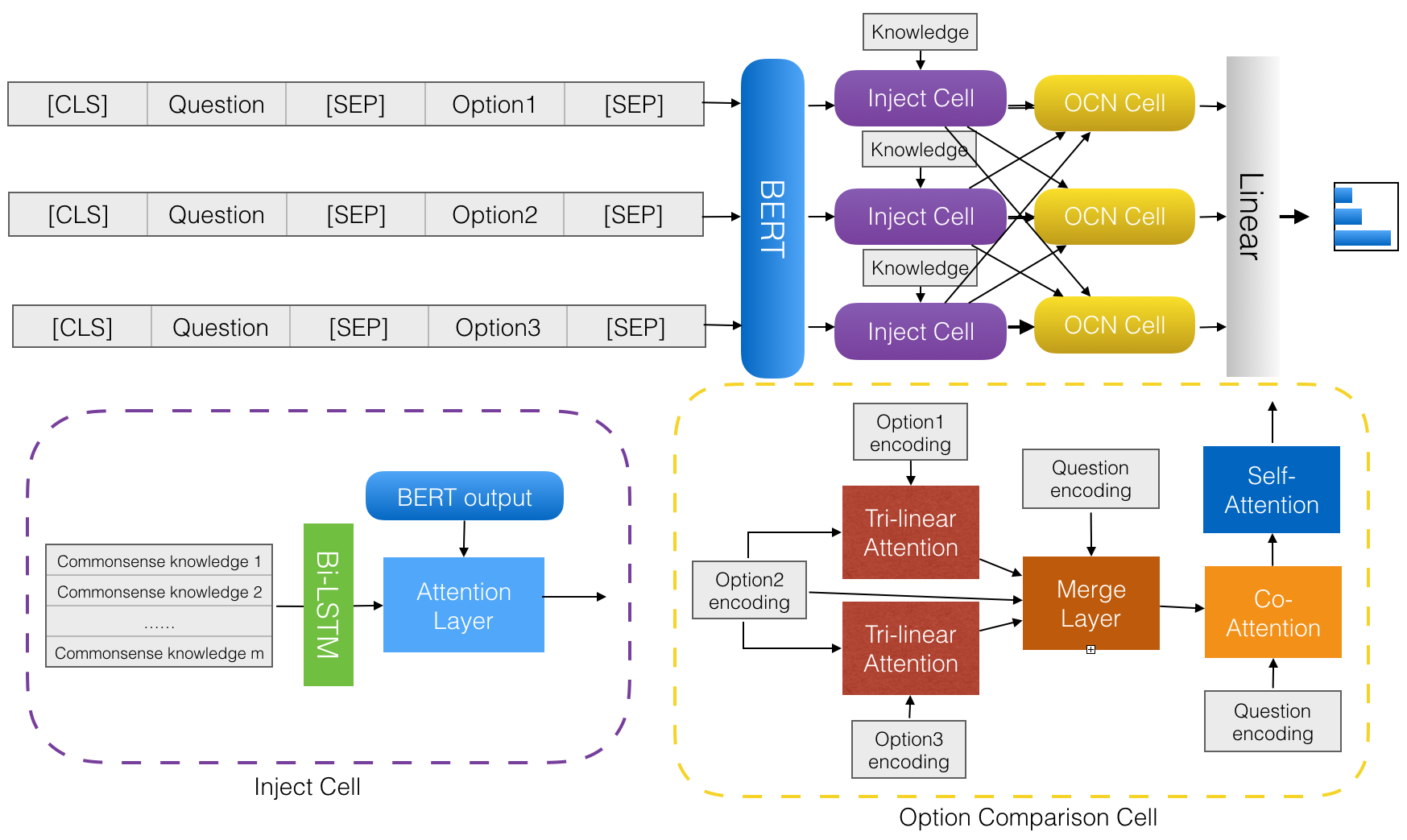}
    \caption{Option Comparison Network with Knowledge Injection}
    \label{fig:model}
\end{figure*}

\subsubsection{Dataset}
\label{datasets}

\texttt{CommonsenseQA} is a multiple-choice QA dataset that specifically measure commonsense reasoning \cite{talmor-etal-2019-commonsenseqa}. This dataset is constructed based on \texttt{ConceptNet} (see section \ref{kb} for more information about this knowledge base). Specifically, a source concept is first extracted from \texttt{ConceptNet}, along with 3 target concepts that are connected to the source concept, i.e., a sub-graph. Crowd-workers are then asked to generate questions, using the source concept, such that only one of the target concepts can correctly answer the question. Additionally, 2 more ``distractor'' concepts are selected by crowd-workers, so that each question is associated with 5 answer-options. In total, the dataset contains 12,247 questions. For \texttt{CommonsenseQA}, we evaluate models on the development-set only, since test-set answers are not publicly available.

\subsubsection{Knowledge bases}
\label{kb}
The first knowledge-base we consider for our experiments is \texttt{ConceptNet} \cite{liu2004conceptnet}. \texttt{ConceptNet} contains over 21 million edges and 8 million nodes (1.5 million nodes in the partition for the English vocabulary), from which one may generate triples of the form $(C1, r, C2)$, wherein the natural-language concepts $C1$ and $C2$ are associated by commonsense relation $r$, e.g., \textit{(dinner, AtLocation, restaurant)}. Thanks to its coverage, \texttt{ConceptNet} is one of the most popular semantic networks for commonsense. \texttt{ATOMIC} \cite{sap2019atomic} is a knowledge-base that focuses on procedural knowledge. Triples are of the form \textit{(Event, r, \{Effect$|$Persona$|$Mental-state\}}), where head and tail are short sentences or verb phrases and $r$ represents an \textit{if-then} relation type: \textit{(X compliments Y, xIntent, X wants to be nice)}. Since the \texttt{CommonsenseQA} dataset is open-domain and requires general commonsense, we think these knowledge-bases are most appropriate for our investigation.

\subsubsection{Model architecture}

The model class we select is that of the \textit{Bidirectional Encoder Representations with Transformer} (BERT) model \cite{devlin-etal-2019-bert}, as it has been applied to numerous QA tasks and has achieved very promising performance, particularly on the \texttt{CommonsenseQA} dataset. When utilizing BERT on multiple-choice QA tasks, the standard approach is to concatenate the question with each answer-option, in order to generate a list of tokens which is then fed into BERT encoder; a linear layer is added on top, in order to predict the answer. One aspect of this strategy is that each answer-option is encoded independently, which limits the model's ability to find correlations between answer-options and with respect to the original question context. To address this issue, the \textit{Option Comparison Network} (OCN) \cite{DBLP:journals/corr/abs-1903-03033} was introduced to explicitly model the pairwise answer-option interactions, making OCN better-suited for multiple-choice QA task structures. The OCN model uses BERT as its base encoder: the question/option encoding is produced by BERT and further processed in a Option Comparison Cell, before being fed into linear layer. The Option Comparison Cell is illustrated in the bottom right of figure \ref{fig:model}. We re-implemented OCN while keeping BERT as its upstream encoder (we refer an interested reader to \cite{DBLP:journals/corr/abs-1903-03033,ma-etal-2019-towards} for more details).

\subsubsection{Knowledge elicitation}
\textbf{ConceptNet}. We identify \texttt{ConceptNet} relations that connect questions to the answer-options. The intuition is that these relation paths would provide explicit evidence that would help the model find the answer. Formally, given a question $Q$ and an answer-option $O$, we find all \texttt{ConceptNet} relations \textit{(C1, r, C2)}, such that $C1 \in Q$ and $C2 \in O$, or vice versa. This rule works well for single-word concepts. However, a large number of concepts in \texttt{ConceptNet} are actually phrases, where finding exactly matching phrases in $Q/O$ is more challenging. To fully utilize phrase-based relations, we relaxed the exact-match constraint to the following:
\begin{equation}
\frac{\text{\# words in C $\cap$ S}}{\text{\# words in C}} > 0.5 
\end{equation}
Here, the sequence $S$ represents $Q$ or $O$, depending on which sequence we try to match the concept $C$ to. Additionally, when the part-of-speech (POS) tag for a concept is available, we make sure it matches the POS tag of the corresponding word in $Q/O$. Table \ref{csqa-cn} shows the extracted \texttt{ConceptNet} triples for the \texttt{CommonsenseQA} example in Table \ref{csqa-example}. It is worth noting that we are able to extract the original \texttt{ConceptNet} sub-graph that was used to create the question, along with some extra triples. Although not perfect, the bold \texttt{ConceptNet} triple provides clues that could help the model resolve the correct answer.

\begin{table*}[h]
\footnotesize
\begin{center}
\begin{tabular}{|c|l|}
\hline 
\bf Options & \bf Extracted \texttt{ConceptNet} triples \\
\hline
Bank & (revolving door \textit{AtLocation} bank) \bf (bank RelatedTo security) \\
Library & (revolving door \textit{AtLocation} library) \\
Department Store & (revolving door \textit{AtLocation} store) (security IsA department) \\
Mall & (revolving door \textit{AtLocation} mall) \\
New York & (revolving door \textit{AtLocation} New York) \\
\hline
\end{tabular}
\end{center}
\caption{Extracted \texttt{ConceptNet} relations for sample shown in Table \ref{csqa-example}.}
\label{csqa-cn}
\end{table*}

\noindent
\textbf{ATOMIC}. We observe that many questions in the \texttt{CommonsenseQA} task ask about which event is likely to occur, given a condition. Superficially, this particular question type seems well-suited for \texttt{ATOMIC}, whose focus is on procedural knowledge. Thus, we could frame our goal as evaluating whether \texttt{ATOMIC} can provide relevant knowledge to help answer these questions. However, one challenge of extracting knowledge from this resource is that heads and tails of knowledge triples in \texttt{ATOMIC} are short sentences or verb phrases, while rare words and person-references are reduced to blanks and PersonX/PersonY, respectively. 

\subsubsection{Knowledge injection}
Given previously-extracted knowledge triples, we need to integrate them with the OCN component of our model. Inspired by \cite{bauer-etal-2018-commonsense}, we propose to use attention-based injection. For \texttt{ConceptNet} knowledge triples, we first convert concept-relation entities into tokens from our lexicon, in order to generate a pseudo-sentence. For example, ``\textit{(book, AtLocation, library)}'' would be converted to ``book at location library.'' Next, we used the knowledge injection cell to fuse the commonsense knowledge into BERT's output, before feeding the fused output into the OCN cell. Specifically, in a knowledge-injection cell, a Bi-LSTM layer is used to encode these pseudo-sentences, before computing the attention with respect to BERT output, as illustrated in bottom left of figure \ref{fig:model}. 

\subsubsection{Knowledge pre-training}
\label{pretrain}
Pre-training large-capacity models (e.g., BERT, GPT \cite{radford2019language}, XLNet \cite{yang2019xlnet}) on large corpora, then fine-tuning on more domain-specific information, has led to performance improvements on various tasks. Inspired by this, our goal in this section is to observe the effect of pre-training BERT on commonsense knowledge and refining the model on task-specific content from the \texttt{CommonsenseQA} dataset. Essentially, we would like to test if pre-training on our external knowledge resources can help the model acquire commonsense. For the \texttt{ConceptNet} pre-training procedure, pre-training BERT on pseudo-sentences formulated from \texttt{ConceptNet} knowledge triples does not provide much gain on performance. Instead, we trained BERT on the \textit{Open Mind Common Sense} (OMCS) corpus \cite{Singh:2002:OMC:646748.701499}, the originating corpus that was used to create \texttt{ConceptNet}. We extracted about 930K English sentences from OMCS and randomly masked out 15\% of the tokens; we then fine-tuned BERT, using a masked language model objective, where the model's objective is to predict the masked tokens, as a probability distribution over the entire lexicon. Finally, we load this fine-tuned model into OCN framework proceed with the downstream \texttt{CommonsenseQA} task. As for pre-training on \texttt{ATOMIC}, we follow previous work's pre-processing steps to convert \texttt{ATOMIC} knowledge triples into sentences \cite{bosselut-etal-2019-comet}; we created special tokens for 9 types of relations as well as blanks. Next, we randomly masked out 15\% of the tokens, only masking out tail-tokens; we used the same OMCS pre-training procedure. 

\begin{table}[h]
\footnotesize
\begin{center}
\begin{tabular}{|c|c|}
\hline \bf Models & \bf Dev Acc  \\ \hline
BERT + OMCS pre-train(*) & 68.8 \\
RoBERTa + CSPT(*) & \bf 76.2 \\ \hline
OCN & 64.1 \\
OCN + CN injection & 67.3 \\
OCN + OMCS pre-train & 65.2 \\
OCN + ATOMIC pre-train & 61.2 \\
OCN + OMCS pre-train + CN inject & \bf 69.0 \\
\hline
\end{tabular}
\end{center}
\caption{Results on \texttt{CommonsenseQA}; the asterisk (*) denotes results taken from leaderboard.}
\label{csqa-results}
\end{table}


\subsubsection{Results}
For all of our experiments, we run 3 trials with different random seeds and we report average scores tables \ref{csqa-results} and \ref{csqa-errors1}. Evaluated on \texttt{CommonsenseQA}, \texttt{ConceptNet} knowledge-injection provides a significant performance boost (+2.8\%), compared to the OCN baseline, suggesting that explicit links from question to answer-options help the model find the correct answer. Pre-training on OMCS also provides a small performance boost to the OCN baseline.
Since both \texttt{ConceptNet} knowledge-injection and OMCS pre-training are helpful, we combine both approaches with OCN and we are able to achieve further improvement (+4.9\%). Finally, to our surprise, OCN pre-trained on \texttt{ATOMIC} yields a significantly lower performance.

\begin{table*}[h]
\footnotesize
\begin{center}
\adjustbox{max width=\textwidth}{
\begin{tabular}{|c|cccccccc|}
\hline \bf Models & \bf AtLoc.(596) & \bf Cau.(194) & \bf Cap.(109) & \bf Ant.(92) & \bf H.Pre.(46) & \bf H.Sub.(39) & \bf C.Des.(28) & \bf Des.(27) \\ \hline
OCN & 64.9 & 66.5 & 65.1 & 55.4 & 69.6 & 64.1 & 57.1 & 66.7\\
+CN inj, & 67.4(+2.5) & 70.6(+4.1) & 66.1(+1.0) & 60.9(+5.5) & 73.9(+4.3)  & 66.7(+2.6) & 64.3(+7.2) & 77.8(+11.1)\\
+OMCS & 68.8(+3.9) & 63.9(-2.6) & 62.4(-2.7) & 60.9(+5.5) & 71.7(+2.1)  & 59.0(-5.1) & 64.3(+7.2) & 74.1(+7.4)\\
+ATOMIC & 62.8(-2.1) & 66.0(\textbf{-0.5}) & 60.6(-4.5) & 52.2(-3.2) & 63.0(-6.6)  & 56.4(-7.7) & 60.7(\textbf{+3.6}) & 74.1(\textbf{+7.4}) \\
+OMCS+CN & 71.6(+6.7) & 71.6(+5.1) & 64.2(+0.9) & 59.8(+4.4) & 69.6(+0.0)  & 69.2(+5.1) & 75.0(+17.9) & 70.4(+3.7) \\
\hline
\end{tabular}}
\end{center}
\caption{Accuracies for each \texttt{CommonsenseQA} question type: \textbf{AtLoc.} means \textit{AtLocation}, \textbf{Cau.} means Causes, \textbf{Cap.} means \textit{CapableOf}, \textbf{Ant.} means \textit{Antonym}, \textbf{H.Pre.} means \textit{HasPrerequiste}, \textbf{H.Sub} means \textit{HasSubevent}, \textbf{C.Des.} means \textit{CausesDesire}, and \textbf{Des.} means \textit{Desires}. Numbers beside types denote the number of questions of that type.}
\label{csqa-errors1}
\end{table*}

\subsubsection{Error Analysis}
To better understand when a model performs better or worse with knowledge-injection, we analyzed model predictions by question type. Since all questions in \texttt{CommonsenseQA} require commonsense reasoning, we classify questions based on the \texttt{ConceptNet} relation between the question concept and correct answer concept. The intuition is that the model needs to capture this relation in order to answer the question. The accuracies for each question type are shown in Table \ref{csqa-errors1}. Note that the number of samples by question type is very imbalanced. Thus due to the limited space, we omitted the long tail of the distribution (about 7\% of all samples). We can see that with \texttt{ConceptNet} relation-injection, all question types got performance boosts|for both the OCN model and OCN model that was pre-trained on OMCS|suggesting that external knowledge is indeed helpful for the task. In the case of OCN pre-trained on \texttt{ATOMIC}, although the overall performance is much lower than the OCN baseline, it is interesting to see that performance for the ``Causes'' type is not significantly affected. Moreover, performance for ``CausesDesire'' and ``Desires'' types actually got much better. As noted by \cite{sap2019atomic}, the ``Causes'' relation in \texttt{ConceptNet} is similar to ``Effects'' and ``Reactions'' in \texttt{ATOMIC}; and ``CausesDesire'' in \texttt{ConceptNet} is similar to ``Wants'' in \texttt{ATOMIC}. This result suggests that models with knowledge pre-training perform better on questions that fit the knowledge domain, but perform worse on others. In this case, pre-training on \texttt{ATOMIC} helps the model do better on questions that are similar to \texttt{ATOMIC} relations, even though overall performance is inferior. Finally, we noticed that questions of type ``Antonym'' appear to be the hardest ones. Many questions that fall into this category contain negations, and we hypothesize that the models still lack the ability to reason over negation sentences, suggesting another direction for future improvement. 

\subsubsection{Discussion}
Based on our experimental results and error analysis, we see that external knowledge is only helpful when there is alignment between questions and knowledge-base types. Thus, it is crucial to identify the question type and apply the best-suited knowledge. In terms of knowledge-injection methods, attention-based injection seems to be the better choice for pre-trained language models such as BERT. Even when alignment between knowledge-base and dataset is sub-optimal, the performance would not degrade. On the other hand, pre-training on knowledge-bases would shift the language model's weight distribution toward its own domain, greatly. If the task domain does not fit knowledge-base well, model performance is likely to drop. When the domain of the knowledge-base aligns with that of the dataset perfectly, both knowledge-injection methods bring performance boosts and a combination of them could bring further gain.

We have presented a survey on two popular knowledge bases (\texttt{ConceptNet} and \texttt{ATOMIC}) and recent knowledge-injection methods (attention and pre-training), on the \texttt{CommonsenseQA} task. We believe it is worth conducting a more comprehensive study of datasets and knowledge-bases and putting more effort towards defining an auxiliary neural learning objectives, in a multi-task learning framework, that classifies the type of knowledge required, based on data characteristics. In parallel, we are also interested in building a \textit{global commonsense knowledge base} by aggregating \texttt{ConceptNet}, \texttt{ATOMIC}, and potentially other resources like FrameNet \cite{baker1998berkeley} and MetaNet \cite{dodge2015metanet}, on the basis of a shared-reference ontology (following the approaches described in \cite{gangemi2010interfacing} and \cite{scheffczyk2010reasoning}): the goal would be to assess whether injecting knowledge structures from a semantically-cohesive lexical knowledge base of commonsense would guarantee stable model accuracy across datasets. 

\section{Conclusion}

We illustrated two projects on computational context understanding through neuro-symbolism. The first project (section \ref{application-1}) concerned the use of knowledge graphs to learning an embedding space for characterising visual scenes, in the context of autonomous driving. The second application (section \ref{application-2}) focused on the extraction and integration of knowledge, encoded in commonsense knowledge bases, for guiding the learning process of neural language models in question-answering tasks. Although diverse in scope and breadth, both projects adopt a hybrid approach to building AI systems, where deep neural networks are enhanced with knowledge graphs. For instance, in the first project we demonstrated that scenes that are visually different can be discovered as sharing similar semantic characteristics by using knowledge graph embeddings; in the second project we showed that a language model is more accurate when it includes specialized modules to evaluate questions and candidate answers on the basis of a common knowledge graph. In both cases, {\it explainability} emerges as a property of the mechanisms that we implemented, through this combination of data-driven algorithms with the relevant knowledge resources. 

We began the chapter by alluding to the way in which humans leverage a complex array of cognitive processes, in order to understand the environment; we further stated that one of the greatest challenges in AI research is learning how to endow machines with similar sense-making capabilities. In these final remarks, it is important to emphasize again (see \textit{footnote \#3}) that the capability we describe here need only follow from satisfying the functional requirements of context understanding, rather than concerning ourselves with how those requirements are specifically implemented in humans versus machines. In other words, our hybrid AI approach stems from the complementary nature of perception and knowledge, but does not commit to the notion of replicating human cognition in the machine: as knowledge graphs can only capture a stripped-down representation of what we know, deep neural networks can only approximate how we perceive the world and learn from it. Certainly, human knowledge (encoded in machine-consumable format) abounds in the digital world, and our work shows that these knowledge bases can be used to instruct ML models and, ultimately, enhance AI systems. 


\bibliography{33-oltramari.bib}
\bibliographystyle{acm.bst}

\end{document}